\documentclass[runningheads]{llncs}
\usepackage[isbn=false]{biblatex}
\AtEveryBibitem{\clearfield{issn}}
\usepackage[T1]{fontenc}
\usepackage{appendix}
\usepackage{enumitem}
\usepackage{eurosym}
\usepackage{graphicx}
\usepackage{subfig}
\usepackage{todonotes}
\usepackage{multirow}
\usepackage{booktabs}
\usepackage{longtable}
\usepackage{amsmath}
\usepackage{hyperref}
\usepackage{makecell}
\usepackage{xcolor}
\usepackage{listings}
\usepackage{float}
\usepackage[most]{tcolorbox}
\newif\ifanon

\ifanon
  \hypersetup{
    pdfauthor={},
    pdftitle={},
    pdfsubject={},
    pdfkeywords={},
    pdfcreator={},
    pdfproducer={}
  }
\fi

\usepackage{graphicx}
\begin{document}

\title{Towards Cost-Efficient Video Dataset Construction with Large Language Models}
\titlerunning{Cost-efficient video dataset construction with LLMs}

%
%

\title{Towards Cost-Efficient Video Dataset Construction with Large Language Models}

\ifanon
  \author{Anonymous Author(s)}
  \authorrunning{Anonymous Author(s)}
  \institute{Institution(s) and address(es) withheld for review}
\else
  \author{Shabot Mohammed\inst{1}\orcidID{0009-0009-5336-1970} \and
          Jonathan Adams\inst{2}\orcidID{0009-0005-2848-438X} \and
          Jan Van den Stock\inst{2}\orcidID{0000-0001-5756-3195} \and
          Joost Vennekens\inst{1}\orcidID{0000-0002-0791-0176}}
  \authorrunning{S. Mohammed et al.}
  \institute{AI Lab, Vrije Universiteit Brussel, Pleinlaan 2, 1050 Brussels, Belgium\\
    \email{mohammed.shabot@vub.be, joost.vennekens@vub.be}\and
    Neuropsychiatry, KU Leuven, ON5/B Herestraat 49 - box 1029, 3000 Leuven, Belgium\\
    \email{jonathan.adams@kuleuven.be, jan.vandenstock@kuleuven.be}}
\fi
\maketitle              

\begin{abstract}
Research on video classification models has witnessed significant growth recently, accompanied by a significant increase in the demand for datasets to train such models. Existing video dataset curation approaches share a common bottleneck: candidate videos must be downloaded before any relevance judgment is made, which results in storage, time and annotation costs. In this work, we present a filtering step, using Large Language Models (LLMs) to pre-screen videos based on their metadata, before actually downloading these videos, thereby saving on these costs. We evaluate the LLMs against a strict human majority vote derived from four independent annotators and validate the reliability of the annotations through inter-annotator agreement analysis (Gwet's AC1). We compare the LLM pre-filter against two no-LLM baselines, namely two keyword heuristics. Via a Leave-One-Rater-Out (LOO) mechanism and an Add-One-Rater-In (AORI), we evaluate the performance of the LLMs as virtual annotators. Finally, we report cost comparisons relative to existing approaches. To ensure reproducability, we made our implementation available on github. \footnote{\url{https://github.com/MoShabot-PhD/llm-paper}}

\end{abstract}
\section{Introduction}

Deep learning methods are now state-of-the-art for many machine learning tasks. One of the main challenges in deep learning lies in the (re-)training of models and the large datasets that are needed for this. This is particularly the case for tasks involving video data, such as action recognition, motion detection or visual question answering. There is a vast amount of video data available on the internet. Some authors claim that approximately 24 hours of video content is uploaded to YouTube every minute \cite{kuehne_hmdb_2011}. The problem is therefore not the availability of the data, but rather the process of constructing a relevant, high-quality dataset.
A common approach to collect video data is to download all candidate videos and manually annotate them \cite{soomro_ucf101_nodate}, possibly with some basic filtering steps such as keyword matching on titles from the YouTube corpus \cite{kay_kinetics_2017}. However, this requires downloading large volumes of content before any relevance judgment can be made, resulting in a significant loss of computational resources, storage and time.
A more recent approach in constructing datasets is using state-of-the-art video generation models. These models allow researchers to construct automated video generation pipelines, in which customizable and detailed prompts can be used to produce video datasets at scale \cite{rahman_all_2026}. This eliminates the need for manual annotation and data scraping, and enables experimental control and customizability of the videos. This process, however, also comes with its own downsides, such as long processing times, video generation costs and potential video model biases being incorporated in the dataset. Moreover, for complex cognitive tasks, such as social cognition, it is not clear if human annotators always interpret generated videos in the same way as natural videos.

In this paper, we propose a novel method of constructing video datasets, by using LLMs to pre-screen the metadata of videos, before downloading them. We show the effectiveness of this method by comparing state-of-the-art LLMs, namely (at the time of writing) GPT-5.2, Claude Sonnet 4.6 and open-source local models: QWEN3.6-36B and Gemma4:31B, against manual annotations at both metadata and video level. We additionally compare the LLMs against two commonly used, no-LLM baselines, to establish whether the LLM pre-filter earns its cost relative to methods a researcher could build for free. We assess the resulting cost reduction relative to traditional pipelines and synthetic video generation pipelines.
Specifically, this work answers the following research questions:
\begin{enumerate}
    \item[\textbf{RQ1}:] Can LLMs be integrated into a scraping pipeline as a high-recall filter on metadata?
    \item[\textbf{RQ2}:] Is integrating LLMs into a scraping pipeline really cost effective?
    \item[\textbf{RQ3}:] Can LLMs substitute for human annotators, and does this depend on the annotator's level of expertise?
\end{enumerate}

\section{Related Work}
In this section, we review a number of existing video datasets, focusing on their construction pipelines and the bottlenecks they share.
Early datasets, such as the UCF101 dataset \cite{soomro_ucf101_nodate} relied on downloading candidate videos and manually removing irrelevant ones, which was common practice at the time. The Human Motion Database was built using a similar approach, scraping videos from several sources (YouTube, Prelinger archive, ...) and assigning students to annotate those manually \cite{kuehne_hmdb_2011}. The construction of the Kinetics dataset started from a list of actions gathered from existing datasets, such as ActivityNet \cite{heilbron_activitynet_2015}, ACT \cite{wang_actions_2016}, MPII Human Pose \cite{andriluka_2d_2014} and the datasets mentioned earlier. Then, videos were chosen by keyword matching the entries in the action list with titles in the YouTube corpus and annotating them by crowd-sourcing the annotation work to Amazon Mechanical Turkers \cite{kay_kinetics_2017}. For the construction of the Something-Something V2 dataset, researchers hired crowd workers to record videos themselves, given a set of predefined labels. The videos were then manually reviewed by Amazon Mechanical Turkers \cite{goyal_something_2017}.
The dataset that most closely aligns with our work, is the recent instruction-following VideoMarathon dataset \cite{lin_unleashing_2025}. The authors collected videos from several publicly available datasets, that covered a diverse range of topics. The videos were annotated by an automated annotation pipeline using QWEN2VL-7B and DeepSeek-v3 \cite{wang_qwen2-vl_nodate, deepseek-ai_deepseek-v3_nodate, lin_unleashing_2025}, resulting in a large, high-quality dataset. This study shows the capabilities of state-of-the-art (open-source, local) video processing models. 
Across all these approaches, a common bottleneck remains: videos must be downloaded before any relevance judgment can be made.

The emerging topic of synthetic data generation avoids this bottleneck entirely, by leveraging the improvement of video generation models. Recent work establishes an automated pipeline to generate a synthetic video dataset, based on starting images and an LLM-generated description of the possible future with respect to the starting frame \cite{rahman_all_2026}. While this avoids the bottleneck, the dataset will inherit limitations from the generator itself.

Both \cite{gu_large_2026} and \cite{haq_llms_2025} have investigated the use of LLMs as virtual annotators, with \cite{gu_large_2026} comparing techniques such as retrieval-augmented generation (RAG) against in-context learning (ICL), and \cite{haq_llms_2025} evaluating LLMs within a full event-annotation workflow. However, these studies primarily evaluate LLMs in relatively controlled annotation settings. In contrast, our work integrates LLMs into a more complex video dataset construction pipeline, where they are used to assess the relevance of candidate videos as part of a multi-stage process involving video retrieval, filtering, and processing. This setting introduces additional challenges, as the LLM must make relevance judgments based on noisy  video metadata rather than well-defined annotation inputs. Our work, therefore, investigates not only the annotation capabilities of LLMs, but also their ability to function as an effective filtering component within a more complex, large-scale dataset construction pipeline.

\section{Method for dataset construction}
In this work, we propose a novel way of constructing video datasets. Instead of downloading all candidate videos before any relevance judgment is made, we propose to have an LLM filter videos based on their metadata alone, avoiding unnecessary downloads, loss of time and downstream annotation costs or video generation costs.
The LLM is tasked with deciding whether a video should be labeled \textit{KEEP} or \textit{DISCARD}, based on the metadata of the video, given explicit instructions through a prompt (Appendix A). Among the available metadata fields, \textit{title} and \textit{description} are passed to the model. To scale this approach, a batch API is used, allowing multiple asynchronous requests to be submitted simultaneously at a cost reduction of 50\%  relative to individual API calls. In addition, the model is prompted to return a confidence score alongside each label. This confidence score is converted into a keep score, equal to the confidence itself if the decision is KEEP, or one minus the confidence if the decision is DISCARD, and the final decision is labeled KEEP only if this keep score exceeds a threshold \textit{t}. The parameter \textit{t} is selected as the F1-maximizing operating point within a single-level GroupKFold procedure (no model is trained, so no inner cross-validation is required): for each outer fold, \textit{t} is determined on the remaining search queries and then applied to the held-out fold, so that no item is evaluated under a threshold tuned on itself. On new dataset domains, \textit{t} could be initialized to 0.5 and tuned when a small subset of labeled data is available. Appendix F shows the (un)tuned parameter \textit{t} and the performance differences.

\section{Concrete use case}
The concrete use case we use in our work is the extension of the FindingEmo image dataset \cite{mertens_findingemo_2024} with video data. FindingEmo is an image dataset containing approximately 25,000 images of complex scenes, where multiple people are interacting in various settings. Images are annotated on several dimensions, such as valence, arousal and emotion label. Annotating this dataset is a demanding task, as there is no objective correct answer: annotators have to interpret a social scene, assess multiple people at once and commit to a judgment on a continuous scale (valence, arousal) and a discrete emotion scale. Extending the image dataset with videos adds other layers of complexity. Video data must meet a number of criteria: the clips should show clear social interaction, contain no on-screen text, and not be documentaries or music videos, just to name a few.

To extend the dataset with video material, we are developing a multi-stage filtering pipeline in which an LLM serves as a pre-filter on metadata. This pre-filtering step is the focus of this present work. 

\section{Experimental setup}
To evaluate the performance of the LLMs, we set up a human annotation panel.
The \emph{baseline panel} consists of four independent human annotators with varying levels of expertise: two domain experts and two non-experts who annotated solely on the basis of written instructions as used in the LLM prompt, shown in Appendix A. Here, the non-experts serve as a proxy for the crowd workers one would recruit on platforms such as Prolific, who rely on written instructions rather than domain knowledge. Comparing the LLMs against the non-experts addresses whether the LLMs can substitute for crowdsourced annotations. The comparison against the experts shows whether the LLMs can approximate expert judgment.
For each LLM, four \emph{mixed panels} are constructed by substituting each human annotator in turn with the LLM. Then, for each panel, we also add the LLM as a fifth annotator and assess whether the panel-agreement significantly degrades or not.

For each panel, we calculate inter-annotator agreement using Gwet's AC1
\cite{gwet_handbook_2001, feinstein_high_1990}. Panel-level AC1 is defined
as the mean of the $\binom{4}{2}=6$ pairwise AC1 scores. For a rater pair
$(a,b)$, pairwise AC1 is computed as
\begin{align}
  \bar{p}_{\text{keep}} &= \tfrac{1}{2}\!\left(
      \Pr(a{=}\textsc{Keep}) + \Pr(b{=}\textsc{Keep})\right), \notag\\
  P_e  &= 2\,\bar{p}_{\text{keep}}\!\left(1-\bar{p}_{\text{keep}}\right),
      \label{eq:pe}\\
  \mathrm{AC1} &= \frac{P_o - P_e}{1 - P_e}, \label{eq:ac1}
\end{align}
where $P_o = \Pr(a = b)$ is the observed agreement. Equation \ref{eq:pe} depicts the probability that two annotators agree by chance \footnote{Equation~\eqref{eq:pe}: binary case of Gwet's chance-agreement term $P_e = \frac{1}{K-1}\sum_{k=1}^{K}\pi_k(1-\pi_k)$.
For $K=2$ we have $\pi_2 = 1-\pi_1$, which reduces to
$2\bar{p}_{\textsc{keep}}(1-\bar{p}_{\textsc{keep}})$.}.
The reported $\Delta_{\mathrm{LOO}}$ is the difference between the AC1 of the original, human-only panel and a panel in which one of the humans has been replaced by an LLM.
Confidence intervals are obtained via cluster-bootstrap resampling at the
search-query level (1000 iterations, 5\textsuperscript{th}--95\textsuperscript{th}
percentile).
A $\Delta_{\mathrm{LOO}}$ is flagged as a significant degradation
(${\downarrow}$) when the 90\,\% CI lies entirely below zero; otherwise it is treated as non-significant (${\sim}$).

To scrape videos and their metadata, we set up predefined lists of emotions, social settings and age groups, to construct our search queries, just like the authors of the FindingEmo dataset did \cite{mertens_findingemo_2024}. For each search query, only the first 10 results were considered, as these are the most likely candidates a researcher would encounter in practice, given that YouTube's ranking algorithm prioritizes relevance \cite{noauthor_how_nodate}. From those 10 videos, we scraped the following seven metadata fields: \textit{title}, \textit{description}, \textit{duration}, \textit{url}, \textit{view\_count}, \textit{uploader} and \textit{uploader\_date}, but ended up only passing title and description to the LLMs, as these contain semantically valuable information for the final decision. In-depth details on the metadata are shown in Appendix D.

\subsubsection{Annotation procedure and inter-annotator agreement}
All items were annotated by four human annotators, each labeling the metadata with KEEP or DISCARD labels. The ground truth label was determined by a strict majority vote: an item is labeled KEEP when at least three of the four annotators chose KEEP, so 2--2 ties (167 items) are labeled DISCARD. Inter-annotator agreement among the four human annotators was calculated using Gwet's AC1 \cite{gwet_handbook_2001}. We chose Gwet's AC1 as a measurement, over the well known Fleiss $\kappa$ to mitigate the kappa paradox which occurs with imbalanced class distributions \cite{feinstein_high_1990}. Human AC1 is reported as a validity check.

We evaluate on the 1047 items for which all four annotators provided a metadata label and all four LLMs returned a usable decision and confidence score. In addition, 298 items were annotated on video-level.
\subsubsection{Baselines}
To assess the LLMs' performance, we compare all models against two, no-LLM baselines. \textbf{Regex A} applies a stop-list of patterns matching the exclusion criteria stated in the prompt, such as music videos or dance performances. An item is kept unless an entry from the stop-list matches. This mirrors the high-recall instruction given to the LLMs and is the closest to the keyword matching used in the construction of the Kinetics dataset \cite{kay_kinetics_2017}. \textbf{Regex B} requires an emotionally loaded keyword to be present (examples are: "wedding", "crying" or "laughter")

\subsubsection{Statistical comparison}
The differences between the methods are assessed with a paired cluster bootstrap: search queries are resampled with replacement over 1000 iterations and the difference in the metric of interest is computed within each resample. We report the 2.5\textsuperscript{th}--97.5\textsuperscript{th} percentile of the resulting difference distribution and treat a difference as significant when the interval excludes zero. We compare the LLMs against the two no-LLM baselines.

\subsubsection{Leave-One-Rater-Out (LOO) and Add-One-Rater-In (AORI)}
To study the performance of the different LLMs, we chose to 1) replace each annotator, one at a time, with the annotations of the LLM to determine whether this results in significant differences and 2) add the LLM as a 5\textsuperscript{th} annotator. This allows us to determine whether adding an LLM as an annotator degrades or improves the overall performance of the panel.

\subsubsection{Query setup}
We reused the scraping queries from the FindingEmo paper \cite{mertens_findingemo_2024}, which consists of lists of emotions, social settings and age groups, and constructed search queries as the Cartesian products of these lists.

The local models (QWEN3.6-36B and Gemma4:31B) were deployed on an Apple M4 Max with 48 GB RAM, using Ollama with Q4\_K\_M quantization for both models. The API models were queried with temperature 1, the local models with temperature 0.3, top-p 0.9 and a fixed seed.
\section{Results}
\subsubsection{Pairwise Human AC1}
Before looking at the performances of the LLMs, we take a look at the pairwise human AC1s, where A and B are domain experts, and C and D are the non-experts. The results are shown in Table \ref{tab:pairwise-ac1}. These results will be used to compare the performances and to answer RQ3. Notice that there already is discrepancy between the different annotators, as the AC1 ranges from 0.507 (moderate agreement) to 0.694 (substantial agreement), due to the ambiguity of the task. Notice that the agreement between the two domain experts is lower than the agreement between one domain expert and a non-expert, confirming the complexity of the task.

\begin{table}[ht]
\centering
\caption{Pairwise human AC1, metadata panel ($n=1047$).}
\label{tab:pairwise-ac1}
\begin{tabular}{lcccccc}
\toprule
\textbf{Annotator pair} & A -- D & A -- C & A -- B & C -- D & B -- D & B -- C \\
\midrule
\textbf{AC1} & 0.694 & 0.652 & 0.611 & 0.550 & 0.529 & 0.507 \\
\bottomrule
\end{tabular}
\end{table}

Table \ref{tab:metadata_results} shows the precision, recall, and F1 metrics of each model at the metadata level, along with the 95\% confidence intervals computed via cluster resampling over 1000 iterations and shows the no-LLM baselines in the lower block. We opted for cluster resampling because the 10 results within each search query are not independent observations as they share a common retrieval bias imposed by YouTube's ranking algorithm.

\begin{table}[ht]
\caption{Metadata-level classification performance (KEEP as positive class),
$n=1047$, 95\% cluster-bootstrap CIs. Lower block: no-LLM baselines.}
\label{tab:metadata_results}
\centering
\small
\begin{tabular}{lcccccc}
\hline
\textbf{Model} & \textbf{Prec.} & \textbf{95\% CI} & \textbf{Recall}
& \textbf{95\% CI} & \textbf{F1} & \textbf{95\% CI} \\
\hline
QWEN3.6-36B        & \textbf{0.354} & [0.271, 0.434] & 0.565 & [0.480, 0.645]
                   & \textbf{0.435} & [0.357, 0.504] \\
Gemma4:31B         & 0.261 & [0.193, 0.331] & \textbf{0.857} & [0.791, 0.917]
                   & 0.401 & [0.314, 0.478] \\
GPT-5.2            & 0.285 & [0.198, 0.373] & 0.526 & [0.421, 0.621]
                   & 0.370 & [0.276, 0.450] \\
Claude Sonnet 4.6  & 0.286 & [0.211, 0.366] & 0.604 & [0.506, 0.695]
                   & 0.388 & [0.307, 0.464] \\
\hline
Regex B (stoplist+keyword) & \textbf{0.269} & [0.183, 0.353] & 0.390 & [0.286, 0.492]
                   & \textbf{0.318} & [0.231, 0.401] \\
Regex A (stoplist) & 0.181 & [0.127, 0.233] & \textbf{0.708} & [0.624, 0.793]
                   & 0.288 & [0.215, 0.353] \\
\hline
\end{tabular}
\end{table}

Point estimates suggest that QWEN3.6-36B is the most balanced model out of all four models, achieving an F1 score of 0.435 and a Recall score of 0.565. Given that this is a pre-filter, we argue that recall is of importance, which makes Gemma4:31B the most favorable model for the task.

\begin{figure}[t]
\centering
\includegraphics[width=0.45\linewidth]{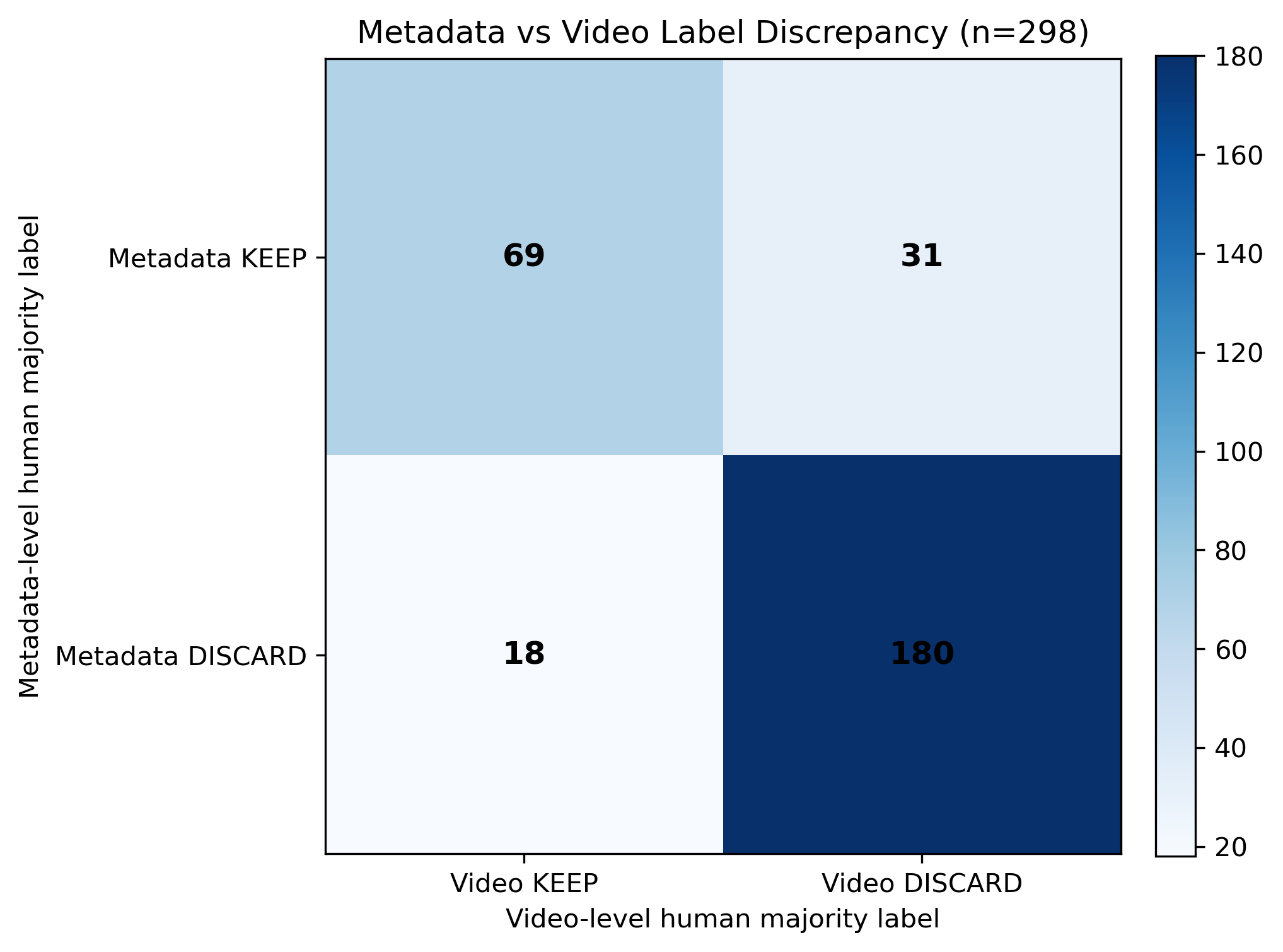}
\caption{Confusion matrix between metadata-level and video-level human majority labels (2-of-3 majority of annotators A, B and C, $n=298$)}. 
\label{fig:meta_vs_video}
\end{figure}

Of the 298 items, there is an 83.6\% agreement between the metadata  labels and the video labels. The 16.4\% disagreement is assymmetric: 31 items labeled KEEP at metadata level were flipped to DISCARD after watching the video, while only 18 flipped the other way. The results for the video level annotations are shown in Figure \ref{fig:meta_vs_video}.

\subsubsection{LOO and AORI}
We assessed the performances of the LLMs by replacing each annotator, one at a time, by the LLM (LOO) and adding each model, one at a time, as a 5\textsuperscript{th} annotator to the panel. Table \ref{tab:loo-metadata} shows the results for the LOO-study and Table \ref{tab:aori} shows the results for the AORI-study.

\begin{table}[ht]
\caption{Leave-one-rater-out, metadata level ($n=1047$): change in panel AC1
($\Delta_{\mathrm{LOO}}$, 90\% CI) when each human annotator is replaced by
an LLM. Reference (4-human) panel AC1 $=0.59$. Every substitution is a
significant degradation (90\% CI entirely below zero).}
\label{tab:loo-metadata}
\centering
\small
\setlength{\tabcolsep}{6pt}
\begin{tabular}{llcc}
\hline
\textbf{Model} & \textbf{Rater} & $\boldsymbol{\Delta_{\mathrm{LOO}}}$ & \textbf{90\% CI} \\
\hline
GPT-5.2 & A & $-0.2973$ & $[-0.3390,\,-0.2552]$ \\
        & B & $-0.2723$ & $[-0.3142,\,-0.2290]$ \\
        & C & \textbf{$-$0.2311} & $[-0.2814,\,-0.1823]$ \\
        & D & $-0.2742$ & $[-0.3177,\,-0.2314]$ \\
\hline
Claude Sonnet 4.6 & A & $-0.3304$ & $[-0.3745,\,-0.2867]$ \\
                  & B & $-0.3036$ & $[-0.3498,\,-0.2626]$ \\
                  & C & \textbf{$-$0.2665} & $[-0.3195,\,-0.2132]$ \\
                  & D & $-0.3107$ & $[-0.3581,\,-0.2655]$ \\
\hline
QWEN3.6-36B & A & $-0.1086$ & $[-0.1291,\,-0.0876]$ \\
            & B & $-0.0438$ & $[-0.0709,\,-0.0186]$ \\
            & C & \textbf{$-$0.0357} & $[-0.0651,\,-0.0065]$ \\
            & D & $-0.0745$ & $[-0.1021,\,-0.0466]$ \\
\hline
Gemma4:31B & A & $-0.2422$ & $[-0.2778,\,-0.2054]$ \\
           & B & $-0.1999$ & $[-0.2382,\,-0.1628]$ \\
           & C & \textbf{$-$0.1679} & $[-0.2126,\,-0.1209]$ \\
           & D & $-0.2195$ & $[-0.2578,\,-0.1795]$ \\
\hline
\end{tabular}
\end{table}

\begin{table}[ht]
\caption{Add-one-rater-in (AORI), metadata level ($n=1047$): change in panel
AC1 ($\Delta_{\mathrm{AORI}}$, 90\% CI) when an LLM is added as a fifth annotator
alongside a four-human panel. Reference panel AC1 $=0.59$.}
\label{tab:aori}
\centering
\small
\begin{tabular}{lccc}
\hline
\textbf{Model} & \textbf{5-rater panel AC1} & \textbf{$\Delta_{\mathrm{AORI}}$} & \textbf{90\% CI} \\
\hline
GPT-5.2            & 0.3755 & $-0.2150$ & [$-0.2503$, $-0.1802$] \\
Claude Sonnet 4.6  & 0.3482 & $-0.2423$ & [$-0.2793$, $-0.2068$] \\
QWEN3.6-36B        & 0.5379 & \textbf{$-$0.0525} & [$-0.0709$, $-0.0350$] \\
Gemma4:31B         & 0.4246 & $-0.1659$ & [$-0.1968$, $-0.1354$] \\
\hline
\end{tabular}
\end{table}

When examining the results, it becomes clear that the substitution of any annotator with an LLM results in a significant degradation of the four-human annotator panel. The degradation which occurs by using QWEN3.6-36B as a substitute is minimal, meaning that QWEN3.6-36B aligns the most with each member of the panel. The same outcome is shown in Table \ref{tab:aori}, where the change in panel AC1 decreases the least (-0.0525) when using QWEN3.6-36B as an extra annotator in the annotator panel. 

\begin{table}[t]
\caption{Full pipeline cost for $n=298$ candidates (\$8/hr, 3 min/video). Local models and baselines incur no-LLM cost.}
\label{tab:cost_comparison_cost}
\centering
\small
\begin{tabular}{l@{\qquad}r@{\qquad}r@{\qquad}r@{\qquad}r}
\hline\noalign{\smallskip}
\multicolumn{1}{c@{\qquad}}{Method} &
\multicolumn{1}{c@{\qquad}}{Total} &
\multicolumn{1}{c@{\qquad}}{Saving} &
\multicolumn{1}{c@{\qquad}}{\makecell{Cost/TP\\(meta)}} &
\multicolumn{1}{c}{\makecell{Cost/TP\\(video)}} \\
\noalign{\smallskip}\hline\noalign{\smallskip}

Manual (no filter)         & \$119.20 & ---         & \$1.19 & \$1.37 \\
GPT-5.2                    & \$47.47  & 2.5$\times$ & \$0.75 & \$0.97 \\
Gemma4:31B                 & \$46.00  & 2.6$\times$ & \textbf{\$0.55} & \textbf{\$0.73} \\
QWEN3.6-36B                & \$48.00  & 2.5$\times$ & \$0.59 & \$0.74 \\
Claude Sonnet 4.6          & \$54.28  & 2.2$\times$ & \$0.66 & \$0.87 \\
Regex A (stoplist)         & \$71.20  & 1.7$\times$ & \$1.02 & \$1.13 \\
Regex B (stoplist+keyword)    & \$27.60  & 4.3$\times$ & \$0.71 & \$0.75 \\

\noalign{\smallskip}\hline
\end{tabular}
\end{table}

To answer RQ2, we calculated the costs using the Batch API, which provides a 50\% discount on the API prices, resulting in \$0.875 per 1M input tokens and \$7.00 per 1M output tokens for GPT-5.2, and \$1.50 per 1M input tokens and \$7.50 per 1M output tokens for Claude Sonnet 4.6. GPT-5.2 used 1,583,031 input tokens and 162,484 output tokens, resulting in a total of 1,745,515 total used tokens, which amounts to $\approx$ \$2.52, whereas Claude Sonnet 4.6 used 1,807,021 input tokens and 132,212 output tokens, resulting in a total token amount of 1,939,233 tokens, which amounts to $\approx$ \$3.70. To determine which model is the most cost effective, we calculated the Cost per True Positive (TP) on the video level annotations. These results are presented in Table \ref{tab:cost_comparison_cost}. We adopt \$8/hour as a crowdsourcing rate, because this is Prolific's recommended minimum compensation for research participants \cite{noauthor_pricing_nodate}. We treat \$8/hour as a conservative estimate. Out of all models, Claude Sonnet 4.6 is the most expensive one, requiring almost \$1 for a relevant video, whereas QWEN3.6-36B and Gemma4:31B are the cheapest options.
Recent advances in video generation models have enabled researchers to generate video datasets using carefully curated prompts. Generating a synthetic dataset of 298 5-second videos costs \$29.95 for the starting images (NanoBananaPro, \$0.10052 per image) and \$223.50 for the videos (Seedance 2.0, \$0.15/s), \$253.45 in total. Consequently, using video generation models to generate videos, the cost per video is approximately \$0.85, assuming that every generated clip is usable, not taking into account generation artifacts. Table \ref{tab:cost_comparison_cost} shows that the best metadata filter costs \$0.73 per video, which is cheaper than the cost per video for video generation models, not taking into account the usability of the generated videos.

\section{Limitations}
Some limitations of this work should be acknowledged. First, the video level annotation subset is relatively small, consisting of only 298 videos. While sufficient for a proof-of-concept, the reported metrics may not be fully representative of performance on a larger scale. 
Second, the threshold parameter was tuned after labeling the dataset, which means that the parameter \textit{t} is only usable in this specific setting. 
Lastly, the prompt that was passed to each model was not prompt engineered with respect to each model. Future work could study the difference in performance when prompt engineering or combining each model with a RAG.

\section{Conclusions}
To answer the research questions mentioned at the start of this work, we evaluated each LLM with a Leave-One-Rater-Out mechanism and an Add-One-Rater-In mechanism and calculated the AC1. Results have shown that for all models, the performance declines, meaning that no annotator can be substituted by an LLM. QWEN3.6-36B aligns the most with the annotator panel. This outcome is also shown in the Add-One-Rater-In mechanism, where the AC1 minimally decreases (-0.0525), which means that QWEN3.6-36B comes closest to acting as a 5th annotator, although this decrease is still significant. All of the models were also compared to the regex baselines which are used by other dataset construction procedures. QWEN3.6-36B is more cost effective in comparison to both regex filters and shows better performance overall.

Taken together, the three research questions receive different degrees of support. RQ1 is confirmed: one of the four models retains high recall (0.857) and all four outperform at least one no-LLM baseline on F1 and recall. RQ2 is confirmed on the video-validated subset, where every LLM pipeline costs less per relevant video recovered than unfiltered manual annotation. The same pattern holds on the full sample using metadata-level proxy, though this has not yet been validated against actual video content. RQ3 receives a cautious answer: at metadata level, no model safely substitutes for or joins any human annotator, expert or non-expert, without significantly degrading panel agreement, though QWEN3.6-36B is consistently the least disruptive.
No single model is "best" across these questions. QWEN3.6-36B achieves the highest F1 and is the most stable virtual annotator, but has a lower level of recall, a weakness given our own argument that recall should be prioritized in a first-stage pre-filter, since a relevant video incorrectly discarded is lost from the pipeline entirely. Gemma4:31B better satisfies that recall-first criterion, at the cost of a slightly lower F1. Choosing between the models is therefore a matter of deciding which error a given pipeline can tolerate less: missing a relevant video (favoring Gemma4:31B) or paying to review more irrelevant ones downstream (favoring QWEN3.6-36B). We therefore answer RQ3 as follows: LLMs, in this complex case, cannot yet substitute human annotators at metadata level, irrespective of the annotator's expertise. QWEN3.6-36B comes closest to a viable substitute. This does not mean LLM pre-filtering is without value. Against the alternative that would be realistically used, such as keyword matching, QWEN3.6-36B and Gemma4:31B outperform both regex baselines on F1. Gemma4:31B outperforms regex A on all metrics.
\printbibliography

@misc{gu_large_2026,
	title = {Large {Language} {Models} {Are} {Effective} {Human} {Annotation} {Assistants}, {But} {Not} {Good} {Independent} {Annotators}},
	url = {https://aclanthology.org/2026.findings-acl.4/},
	doi = {10.18653/v1/2026.findings-acl.4},
	author = {Gu, Feng and Li, Zongxia and Colon, Carlos Rafael and Evans, Benjamin and Mondal, Ishani and Boyd-Graber, Jordan Lee},
	month = apr,
	year = {2026},
	note = {arXiv:2503.06778 [cs.CL]},
}

@inproceedings{mertens_findingemo_2024,
	address = {Vancouver, BC, Canada},
	title = {{FindingEmo}: {An} {Image} {Dataset} for {Emotion} {Recognition} in the {Wild}},
	isbn = {979-8-3313-1438-5},
	shorttitle = {{FindingEmo}},
	doi = {10.52202/079017-0161},
	urldate = {2026-08-25},
	booktitle = {Advances in {Neural} {Information} {Processing} {Systems} 37},
	publisher = {Neural Information Processing Systems Foundation, Inc. (NeurIPS)},
	author = {Mertens, Laurent and Yargholi, Elahe' and Beeck, Hans De and Stock, Jan Van Den and Vennekens, Joost},
	year = {2024},
	pages = {4956--4996},
}

@inproceedings{lin_unleashing_2025,
	address = {San Diego, California, USA and Mexico City, Mexico},
	title = {Unleashing {Hour}-{Scale} {Video} {Training} for {Long} {Video}-{Language} {Understanding}},
	isbn = {979-8-3313-3827-5},
	doi = {10.52202/085713-0591},
	booktitle = {Advances in {Neural} {Information} {Processing} {Systems} 38},
	publisher = {Neural Information Processing Systems Foundation, Inc. (NeurIPS)},
	author = {Lin, Jingyang and Wu, Jialian and Sun, Ximeng and Wang, Ze and Liu, Jiang and Su, Yusheng and Yu, Xiaodong and Chen, Hao and Luo, Jiebo and Liu, Zicheng and Barsoum, Emad},
	year = {2025},
	pages = {19950--19979},
}

@inproceedings{wang_actions_2016,
	address = {Las Vegas, NV, USA},
	title = {Actions {\textasciitilde} {Transformations}},
	isbn = {978-1-4673-8851-1},
	doi = {10.1109/CVPR.2016.291},
	booktitle = {2016 {IEEE} {Conference} on {Computer} {Vision} and {Pattern} {Recognition} ({CVPR})},
	publisher = {IEEE},
	author = {Wang, Xiaolong and Farhadi, Ali and Gupta, Abhinav},
	month = jun,
	year = {2016},
	pages = {2658--2667},
}

@misc{goyal_something_2017,
	title = {The "something something" video database for learning and evaluating visual common sense},
	doi = {10.1109/ICCV.2017.622},
	publisher = {arXiv},
	author = {Goyal, Raghav and Kahou, Samira Ebrahimi and Michalski, Vincent and Materzyńska, Joanna and Westphal, Susanne and Kim, Heuna and Haenel, Valentin and Fruend, Ingo and Yianilos, Peter and Mueller-Freitag, Moritz and Hoppe, Florian and Thurau, Christian and Bax, Ingo and Memisevic, Roland},
	month = jun,
	year = {2017},
	note = {arXiv:1706.04261 [cs]},
}

@misc{wang_qwen2-vl_nodate,
	title = {Qwen2-{VL}: {Enhancing} {Vision}-{Language} {Model}'s {Perception} of the {World} at {Any} {Resolution}},
	doi = {10.48550/arXiv.2409.12191},
	publisher = {arXiv},
	author = {Wang, Peng and Bai, Shuai and Tan, Sinan and Wang, Shijie and Fan, Zhihao and Bai, Jinze and Chen, Keqin and Liu, Xuejing and Wang, Jialin and Ge, Wenbin and Fan, Yang and Dang, Kai and Du, Mengfei and Ren, Xuancheng and Men, Rui and Liu, Dayiheng and Zhou, Chang and Zhou, Jingren and Lin, Junyang},
	note = {arXiv:2409.12191 [cs]},
}

@misc{soomro_ucf101_nodate,
	title = {{UCF101}: {A} {Dataset} of 101 {Human} {Actions} {Classes} {From} {Videos} in {The} {Wild}},
	shorttitle = {{UCF101}},
	doi = {10.48550/arXiv.1212.0402},
	publisher = {arXiv},
	author = {Soomro, Khurram and Zamir, Amir Roshan and Shah, Mubarak},
	note = {arXiv:1212.0402 [cs]},
}

@misc{rahman_all_2026,
	title = {All in {One}: {A} {Unified} {Synthetic} {Data} {Pipeline} for {Multimodal} {Video} {Understanding}},
	doi = {10.48550/arXiv.2604.12335},
	publisher = {arXiv},
	author = {Rahman, Tanzila and Liao, Renjie and Sigal, Leonid},
	month = apr,
	year = {2026},
	note = {arXiv:2604.12335 [cs]},
}

@inproceedings{kuehne_hmdb_2011,
	address = {Barcelona, Spain},
	title = {{HMDB}: {A} large video database for human motion recognition},
	isbn = {978-1-4577-1102-2 978-1-4577-1101-5 978-1-4577-1100-8},
	shorttitle = {{HMDB}},
	doi = {10.1109/ICCV.2011.6126543},
	booktitle = {2011 {International} {Conference} on {Computer} {Vision}},
	publisher = {IEEE},
	author = {Kuehne, H. and Jhuang, H. and Garrote, E. and Poggio, T. and Serre, T.},
	month = nov,
	year = {2011},
	pages = {2556--2563},
}

@misc{kay_kinetics_2017,
	title = {The {Kinetics} {Human} {Action} {Video} {Dataset}},
	doi = {10.48550/arXiv.1705.06950},
	publisher = {arXiv},
	author = {Kay, Will and Carreira, Joao and Simonyan, Karen and Zhang, Brian and Hillier, Chloe and Vijayanarasimhan, Sudheendra and Viola, Fabio and Green, Tim and Back, Trevor and Natsev, Paul and Suleyman, Mustafa and Zisserman, Andrew},
	month = may,
	year = {2017},
	note = {arXiv:1705.06950 [cs]},
}

@misc{noauthor_how_nodate,
	title = {How {YouTube} search works - {YouTube} {Help}},
	url = {https://support.google.com/youtube/answer/16090438?hl=en},
}

@inproceedings{heilbron_activitynet_2015,
	address = {Boston, MA, USA},
	title = {{ActivityNet}: {A} large-scale video benchmark for human activity understanding},
	isbn = {978-1-4673-6964-0},
	shorttitle = {{ActivityNet}},
	doi = {10.1109/CVPR.2015.7298698},
	booktitle = {2015 {IEEE} {Conference} on {Computer} {Vision} and {Pattern} {Recognition} ({CVPR})},
	publisher = {IEEE},
	author = {Heilbron, Fabian Caba and Escorcia, Victor and Ghanem, Bernard and Niebles, Juan Carlos},
	month = jun,
	year = {2015},
	pages = {961--970},
}

@misc{haq_llms_2025,
	title = {{LLMs} as {Data} {Annotators}: {How} {Close} {Are} {We} to {Human} {Performance}},
	shorttitle = {{LLMs} as {Data} {Annotators}},
	doi = {10.48550/arXiv.2504.15022},
	publisher = {arXiv},
	author = {Haq, Muhammad Uzair Ul and Rigoni, Davide and Sperduti, Alessandro},
	month = apr,
	year = {2025},
	note = {arXiv:2504.15022 [cs.CL]},
}

@article{feinstein_high_1990,
	title = {High agreement but low {Kappa}: {I}. the problems of two paradoxes},
	volume = {43},
	copyright = {https://www.elsevier.com/tdm/userlicense/1.0/},
	issn = {08954356},
	shorttitle = {High agreement but low {Kappa}},
	doi = {10.1016/0895-4356(90)90158-L},
	language = {en},
	number = {6},
	journal = {Journal of Clinical Epidemiology},
	author = {Feinstein, Alvan R. and Cicchetti, Domenic V.},
	month = jan,
	year = {1990},
	pages = {543--549},
}

@misc{deepseek-ai_deepseek-v3_nodate,
	title = {{DeepSeek}-{V3} {Technical} {Report}},
	doi = {10.48550/arXiv.2412.19437},
	publisher = {arXiv},
	author = {DeepSeek-AI and Liu, Aixin and Feng, Bei and Xue, Bing and Wang, Bingxuan and Wu, Bochao and Lu, Chengda and Zhao, Chenggang and Deng, Chengqi and Zhang, Chenyu and Ruan, Chong and Dai, Damai and Guo, Daya and Yang, Dejian and Chen, Deli and Ji, Dongjie and Li, Erhang and Lin, Fangyun and Dai, Fucong and Luo, Fuli and Hao, Guangbo and Chen, Guanting and Li, Guowei and Zhang, H. and Bao, Han and Xu, Hanwei and Wang, Haocheng and Zhang, Haowei and Ding, Honghui and Xin, Huajian and Gao, Huazuo and Li, Hui and Qu, Hui and Cai, J. L. and Liang, Jian and Guo, Jianzhong and Ni, Jiaqi and Li, Jiashi and Wang, Jiawei and Chen, Jin and Chen, Jingchang and Yuan, Jingyang and Qiu, Junjie and Li, Junlong and Song, Junxiao and Dong, Kai and Hu, Kai and Gao, Kaige and Guan, Kang and Huang, Kexin and Yu, Kuai and Wang, Lean and Zhang, Lecong and Xu, Lei and Xia, Leyi and Zhao, Liang and Wang, Litong and Zhang, Liyue and Li, Meng and Wang, Miaojun and Zhang, Mingchuan and Zhang, Minghua and Tang, Minghui and Li, Mingming and Tian, Ning and Huang, Panpan and Wang, Peiyi and Zhang, Peng and Wang, Qiancheng and Zhu, Qihao and Chen, Qinyu and Du, Qiushi and Chen, R. J. and Jin, R. L. and Ge, Ruiqi and Zhang, Ruisong and Pan, Ruizhe and Wang, Runji and Xu, Runxin and Zhang, Ruoyu and Chen, Ruyi and Li, S. S. and Lu, Shanghao and Zhou, Shangyan and Chen, Shanhuang and Wu, Shaoqing and Ye, Shengfeng and Ye, Shengfeng and Ma, Shirong and Wang, Shiyu and Zhou, Shuang and Yu, Shuiping and Zhou, Shunfeng and Pan, Shuting and Wang, T. and Yun, Tao and Pei, Tian and Sun, Tianyu and Xiao, W. L. and Zeng, Wangding and Zhao, Wanjia and An, Wei and Liu, Wen and Liang, Wenfeng and Gao, Wenjun and Yu, Wenqin and Zhang, Wentao and Li, X. Q. and Jin, Xiangyue and Wang, Xianzu and Bi, Xiao and Liu, Xiaodong and Wang, Xiaohan and Shen, Xiaojin and Chen, Xiaokang and Zhang, Xiaokang and Chen, Xiaosha and Nie, Xiaotao and Sun, Xiaowen and Wang, Xiaoxiang and Cheng, Xin and Liu, Xin and Xie, Xin and Liu, Xingchao and Yu, Xingkai and Song, Xinnan and Shan, Xinxia and Zhou, Xinyi and Yang, Xinyu and Li, Xinyuan and Su, Xuecheng and Lin, Xuheng and Li, Y. K. and Wang, Y. Q. and Wei, Y. X. and Zhu, Y. X. and Zhang, Yang and Xu, Yanhong and Xu, Yanhong and Huang, Yanping and Li, Yao and Zhao, Yao and Sun, Yaofeng and Li, Yaohui and Wang, Yaohui and Yu, Yi and Zheng, Yi and Zhang, Yichao and Shi, Yifan and Xiong, Yiliang and He, Ying and Tang, Ying and Piao, Yishi and Wang, Yisong and Tan, Yixuan and Ma, Yiyang and Liu, Yiyuan and Guo, Yongqiang and Wu, Yu and Ou, Yuan and Zhu, Yuchen and Wang, Yuduan and Gong, Yue and Zou, Yuheng and He, Yujia and Zha, Yukun and Xiong, Yunfan and Ma, Yunxian and Yan, Yuting and Luo, Yuxiang and You, Yuxiang and Liu, Yuxuan and Zhou, Yuyang and Wu, Z. F. and Ren, Z. Z. and Ren, Zehui and Sha, Zhangli and Fu, Zhe and Xu, Zhean and Huang, Zhen and Zhang, Zhen and Xie, Zhenda and Zhang, Zhengyan and Hao, Zhewen and Gou, Zhibin and Ma, Zhicheng and Yan, Zhigang and Shao, Zhihong and Xu, Zhipeng and Wu, Zhiyu and Zhang, Zhongyu and Li, Zhuoshu and Gu, Zihui and Zhu, Zijia and Liu, Zijun and Li, Zilin and Xie, Ziwei and Song, Ziyang and Gao, Ziyi and Pan, Zizheng},
	note = {arXiv:2412.19437 [cs]},
}

@inproceedings{andriluka_2d_2014,
	address = {Columbus, OH, USA},
	title = {{2D} {Human} {Pose} {Estimation}: {New} {Benchmark} and {State} of the {Art} {Analysis}},
	isbn = {978-1-4799-5118-5},
	shorttitle = {{2D} {Human} {Pose} {Estimation}},
	doi = {10.1109/CVPR.2014.471},
	booktitle = {2014 {IEEE} {Conference} on {Computer} {Vision} and {Pattern} {Recognition}},
	publisher = {IEEE},
	author = {Andriluka, Mykhaylo and Pishchulin, Leonid and Gehler, Peter and Schiele, Bernt},
	month = jun,
	year = {2014},
	pages = {3686--3693},
}

@article{gwet_handbook_2001,
	title = {Handbook of inter-rater reliability},
	journal = {Gaithersburg, MD: STATAXIS Publishing Company},
	author = {Gwet, Kilem},
	year = {2001},
	pages = {223--246},
}

@misc{noauthor_pricing_nodate,
	title = {Pricing},
	url = {https://www.prolific.com/pricing},
	urldate = {2026-05-05},
	journal = {Prolific pricing},
}
\section{Appendices}
\section*{Appendix A: Prompt}

\begin{quote}
\small
 
You are labeling video metadata for the FindingEmo dataset.
 
\medskip
\noindent\textbf{IMPORTANT}
\begin{itemize}
  \item You do NOT see the video.
  \item You must use ONLY the title and description, including hashtags.
  \item This is a FIRST-STAGE HIGH-RECALL PREFILTER, not a final classifier.
\end{itemize}
 
Your job is to remove only videos that are clearly ineligible. If there is any
realistic possibility of real human presence, interpersonal context, or personal
experience, you should KEEP the video.
 
\medskip
\noindent\textbf{CORE DECISION RULE}
\begin{itemize}
  \item DISCARD only when the metadata explicitly states or strongly implies
        the video is outside scope.
  \item Otherwise KEEP.
  \item When uncertain $\rightarrow$ KEEP.
  \item Prefer false positives over false negatives.
\end{itemize}
 
\medskip
\noindent\textbf{CRITICAL INTERPRETATION RULES}
\begin{itemize}
  \item Absence of emotional language is NOT a reason to discard.
  \item Weak or unclear emotion is NOT a reason to discard.
  \item Absence of people must NOT be assumed unless explicitly stated or
        strongly implied.
  \item If human context is plausible $\rightarrow$ KEEP.
\end{itemize}
 
Use this decision procedure:
 
\medskip
\noindent\textbf{STEP 1 --- HARD DISCARD}\\
Return DISCARD only if metadata clearly matches one of these:
\begin{enumerate}
  \item \textbf{NO PEOPLE / NON-HUMAN ONLY}\\
        Explicit or strongly implied:
        \begin{itemize}
          \item ``no people''
          \item nature, scenery, ambience, landscapes
          \item objects only, background visuals
        \end{itemize}
  \item \textbf{AI-GENERATED / SYNTHETIC}\\
        Explicit mentions of:
        \begin{itemize}
          \item AI-generated, artificial, synthetic, deepfake, avatar, animated
        \end{itemize}
  \item \textbf{PURE INSTRUCTIONAL / TECHNICAL}\\
        Clear tutorial, how-to, tips, product demo, or instructional content
        AND no personal or first-person framing.
  \item \textbf{STOCK / COMPILATION / AESTHETIC}\\
        Stock footage, repost compilations, montages, aesthetic visuals without
        human context.
  \item \textbf{NON-HUMAN FORMAT CONTENT}\\
        Gameplay, screen recordings, slides, text-only, graphics-only.
  \item \textbf{PERFORMANCE-ONLY CONTENT}\\
        Music, dance, or performance with no personal, interpersonal, or
        narrative context.
\end{enumerate}
If none of these clearly apply $\rightarrow$ KEEP.
 
\medskip
\noindent\textbf{STEP 2 --- KEEP SIGNALS}\\
KEEP if there is any indication of:
\begin{itemize}
  \item a person or people
  \item first-person language (I, my, we)
  \item relationships (family, friend, partner)
  \item life events, experiences, or stories
  \item emotional language or emojis
  \item vlog or personal framing
  \item any plausible human or social context
\end{itemize}
 
\medskip
\noindent\textbf{STRONG KEEP SIGNALS}
\begin{itemize}
  \item Personal pronouns: I, my, me, we, our
  \item Relationships: mom, dad, friend, girlfriend, boyfriend, family
  \item Situations: breakup, reunion, decision, story, reaction, vlog, experience
  \item Emojis or emotional words: sad, happy, love, scared
\end{itemize}
 
\medskip
\noindent\textbf{EMOTION LIKELIHOOD} (independent from decision)
\begin{itemize}
  \item high = explicit emotional situation or reaction
  \item medium = human or interpersonal context present
  \item low = little emotional evidence but not excluded
  \item unclear = very limited metadata
\end{itemize}
 
\medskip
\noindent\textbf{AI-GENERATED LIKELIHOOD}
\begin{itemize}
  \item high = explicit AI/synthetic indicators
  \item medium = ambiguous
  \item low = no indication
\end{itemize}
 
\medskip
\noindent\textbf{CONFIDENCE}
\begin{itemize}
  \item 0.90--0.98 = clear DISCARD
  \item 0.85--0.95 = clear KEEP with strong signals
  \item 0.70--0.84 = ambiguous but KEEP
\end{itemize}
 
\medskip
\noindent\textbf{OUTPUT}\\
Return ONLY valid JSON:
\begin{verbatim}
{
  "decision": "KEEP" or "DISCARD",
  "confidence": 0.0,
  "emotion_likelihood": "high|medium|low|unclear",
  "ai_generated_likelihood": "low|medium|high",
  "reason": ""
}
\end{verbatim}
 
\noindent\textbf{Reason}
\begin{itemize}
  \item 1--2 sentences
  \item based only on metadata
  \item explain why rule triggered
\end{itemize}
 
\medskip
\noindent\textbf{EXAMPLES}
 
\medskip
\noindent\textbf{Title:} My dad surprised me after 5 years apart\\
\textbf{Description:} Didn't expect this \#family\\
$\rightarrow$ \textbf{KEEP}
 
\medskip
\noindent\textbf{Title:} A day in my life working night shifts\\
\textbf{Description:} it's been hard lately \#vlog\\
$\rightarrow$ \textbf{KEEP}
 
\medskip
\noindent\textbf{Title:} New skincare routine\\
\textbf{Description:} products that changed my life\\
$\rightarrow$ \textbf{KEEP}
 
\medskip
\noindent\textbf{Title:} Camera test with my girlfriend\\
\textbf{Description:} trying the new lens\\
$\rightarrow$ \textbf{KEEP}
 
\medskip
\noindent\textbf{Title:} Moving to a new city alone\\
\textbf{Description:} starting over is scary\\
$\rightarrow$ \textbf{KEEP}
 
\medskip
\noindent\textbf{Title:} Top 10 productivity hacks\\
\textbf{Description:} improve your workflow\\
$\rightarrow$ \textbf{DISCARD}
 
\medskip
\noindent\textbf{Title:} How to wake up at 5am\\
\textbf{Description:} step by step guide\\
$\rightarrow$ \textbf{DISCARD}
 
\medskip
\noindent\textbf{Title:} Relaxing rain sounds\\
\textbf{Description:} no people, just ambience\\
$\rightarrow$ \textbf{DISCARD}
 
\medskip
\noindent\textbf{Title:} Minecraft gameplay\\
\textbf{Description:} episode 12\\
$\rightarrow$ \textbf{DISCARD}
 
\medskip
\noindent\textbf{Title:} AI-generated short film\\
\textbf{Description:} made with AI tools\\
$\rightarrow$ \textbf{DISCARD}
 
\end{quote}

\section*{Appendix B: Examples of disagreement: non-expert - LLM}
\begin{figure}[H]
  \centering
  \begin{minipage}{0.48\textwidth}
    \centering
    \includegraphics[width=\linewidth]{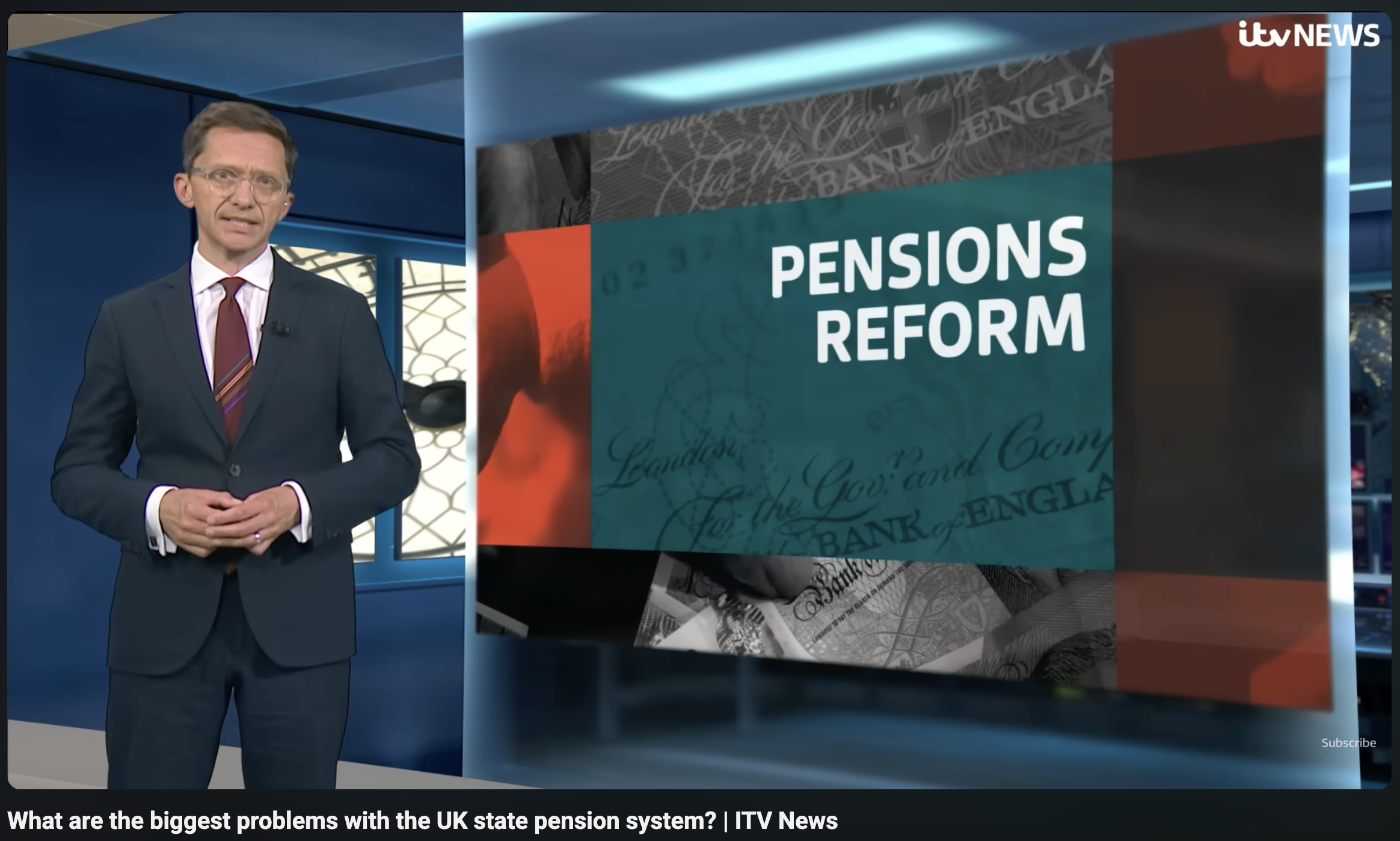}
    \caption{Example of a disagreement where an LLM (GPT in this case) labels DISCARD and the non-expert labels KEEP.
    \newline
    \textbf{link:} \url{https://www.youtube.com/watch?v=K0MG0EAIxPU}}
    \label{fig:LLMDISCARDNEKEEP}
  \end{minipage}\hfill
  \begin{minipage}{0.48\textwidth}
    \centering
    \includegraphics[width=\linewidth]{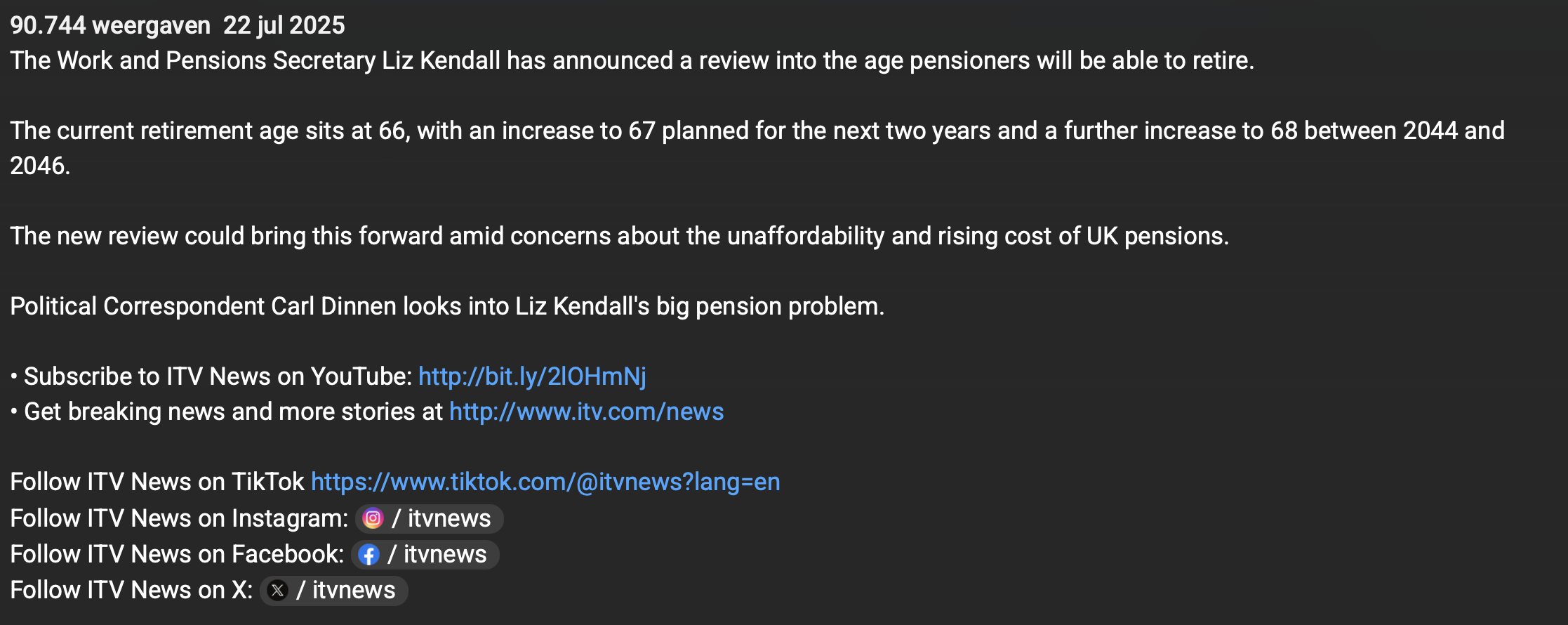}
    \caption{Metadata associated with the video of Figure \ref{fig:LLMDISCARDNEKEEP}}
    \label{fig:LLMDISCARDNEKEEPmeta}
  \end{minipage}
\end{figure}
\begin{figure}[H]
  \centering
  \begin{minipage}{0.48\textwidth}
    \centering
    \includegraphics[width=\linewidth]{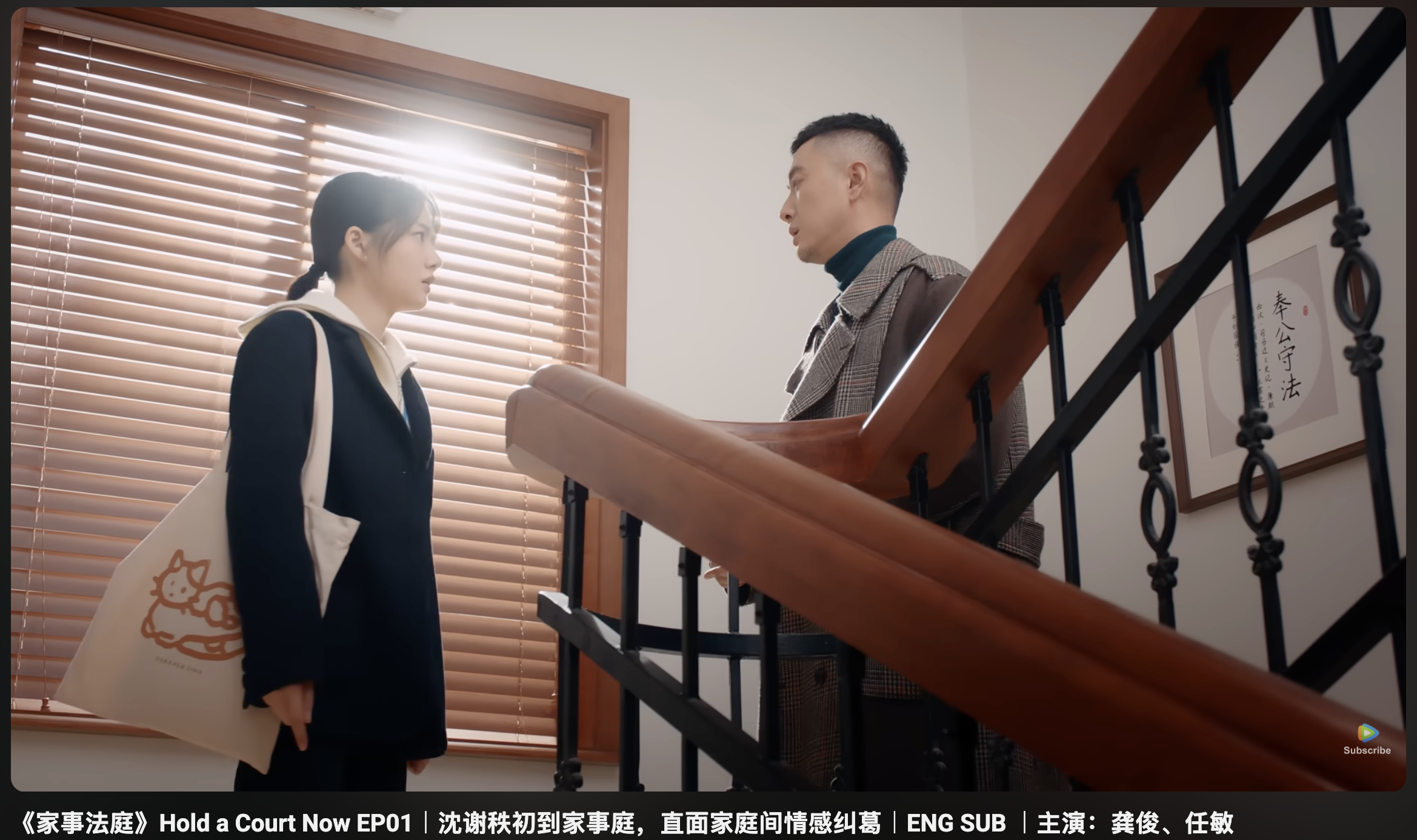}
    \caption{Example of a disagreement where an LLM (GPT in this case) labels KEEP and the non-expert labels DISCARD.
    \newline
    \textbf{link:} \url{https://www.youtube.com/watch?v=FsIDRlJR4Nk}}
    \label{fig:LLMKEEPNEDISCARD}
  \end{minipage}\hfill
  \begin{minipage}{0.48\textwidth}
    \centering
    \includegraphics[width=\linewidth]{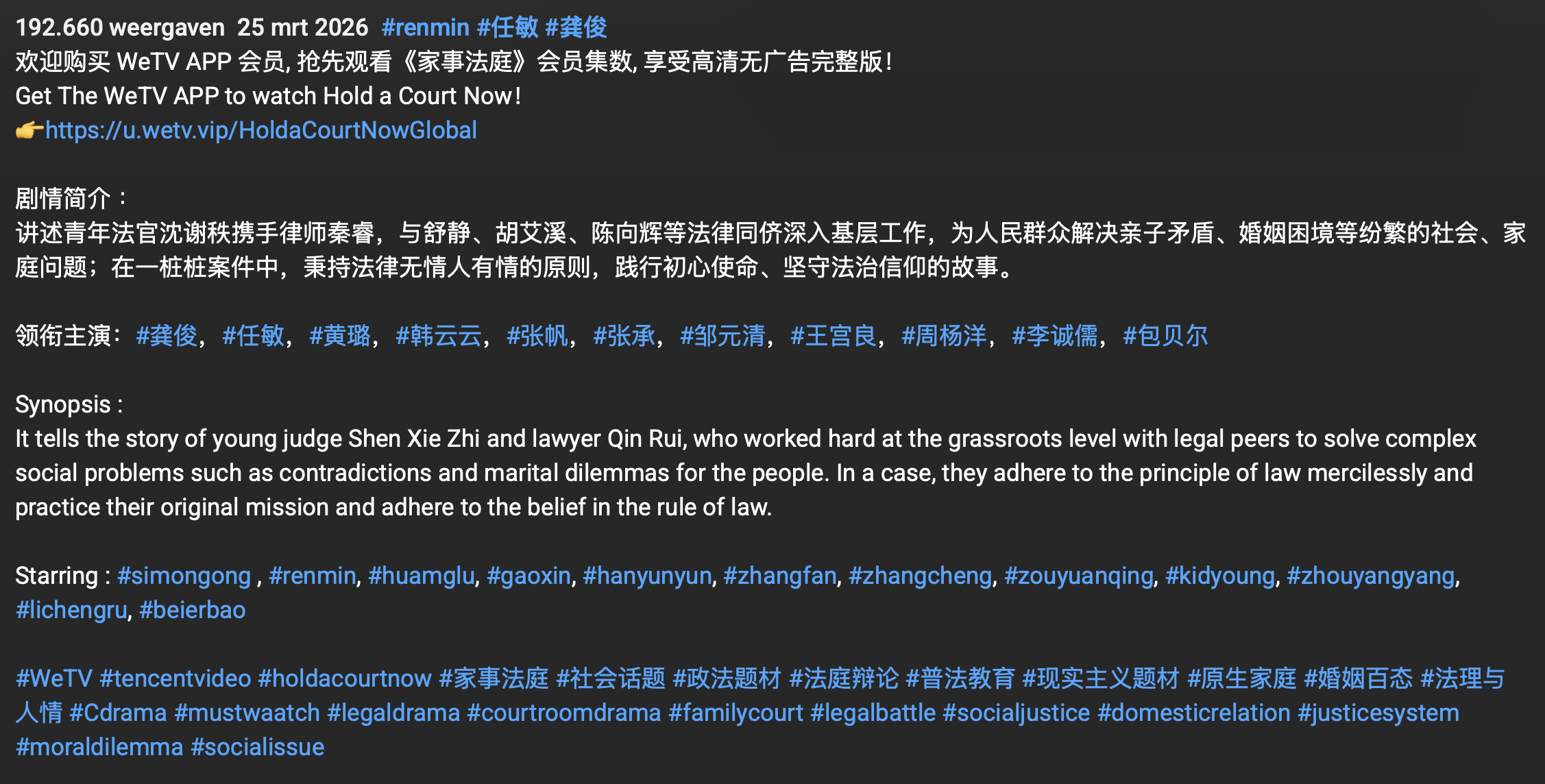}
    \caption{Metadata associated with the video of Figure \ref{fig:LLMKEEPNEDISCARD}}
    \label{fig:LLMKEEPNEDISCARDmeta}
  \end{minipage}
\end{figure}

\section*{Appendix C: Full pairwise comparison matrix}
Exploratory; not corrected for multiplicity. All 14 pairs (4 LLMs, 2 baselines)
compared via paired cluster-bootstrap on metadata-level F1 and recall
($n=1047$, strict human-majority ground truth of the four annotators).

{\footnotesize
\begin{longtable}{llrrc}
\toprule
\textbf{Comparison (A $-$ B)} & \textbf{Metric} & \textbf{$\Delta$} & \textbf{95\% CI} & \textbf{Sig.} \\
\midrule
QWEN3.6-36B -- Regex A (stoplist)            & F1 & $+0.147$ & [$+0.085$, $+0.207$] & yes \\
QWEN3.6-36B -- Regex B (stoplist+keyword)    & F1 & $+0.117$ & [$+0.036$, $+0.198$] & yes \\
Gemma4:31B -- Regex A (stoplist)             & F1 & $+0.112$ & [$+0.065$, $+0.162$] & yes \\
Claude Sonnet 4.6 -- Regex A (stoplist)      & F1 & $+0.100$ & [$+0.037$, $+0.156$] & yes \\
GPT-5.2 -- Regex A (stoplist)                & F1 & $+0.082$ & [$+0.012$, $+0.139$] & yes \\
Gemma4:31B -- Regex B (stoplist+keyword)     & F1 & $+0.082$ & [$+0.013$, $+0.154$] & yes \\
Claude Sonnet 4.6 -- Regex B (stoplist+keyword)& F1 & $+0.070$ & [$-0.003$, $+0.147$] & no \\
GPT-5.2 -- Regex B (stoplist+keyword)        & F1 & $+0.052$ & [$-0.028$, $+0.133$] & no \\
QWEN3.6-36B -- Gemma4:31B                    & F1 & $+0.034$ & [$-0.020$, $+0.088$] & no \\
Claude Sonnet 4.6 -- Gemma4:31B              & F1 & $-0.012$ & [$-0.073$, $+0.045$] & no \\
GPT-5.2 -- Claude Sonnet 4.6                 & F1 & $-0.018$ & [$-0.067$, $+0.030$] & no \\
GPT-5.2 -- Gemma4:31B                        & F1 & $-0.031$ & [$-0.099$, $+0.032$] & no \\
Claude Sonnet 4.6 -- QWEN3.6-36B             & F1 & $-0.047$ & [$-0.123$, $+0.021$] & no \\
GPT-5.2 -- QWEN3.6-36B                       & F1 & $-0.065$ & [$-0.139$, $+0.015$] & no \\
\midrule
Gemma4:31B -- Regex B (stoplist+keyword)     & Recall & $+0.468$ & [$+0.378$, $+0.575$] & yes \\
Claude Sonnet 4.6 -- Regex B (stoplist+keyword)& Recall & $+0.214$ & [$+0.111$, $+0.331$] & yes \\
QWEN3.6-36B -- Regex B (stoplist+keyword)    & Recall & $+0.175$ & [$+0.052$, $+0.293$] & yes \\
Gemma4:31B -- Regex A (stoplist)             & Recall & $+0.149$ & [$+0.058$, $+0.248$] & yes \\
GPT-5.2 -- Regex B (stoplist+keyword)        & Recall & $+0.136$ & [$+0.021$, $+0.262$] & yes \\
Claude Sonnet 4.6 -- QWEN3.6-36B             & Recall & $+0.039$ & [$-0.069$, $+0.159$] & no \\
GPT-5.2 -- QWEN3.6-36B                       & Recall & $-0.039$ & [$-0.164$, $+0.084$] & no \\
GPT-5.2 -- Claude Sonnet 4.6                 & Recall & $-0.078$ & [$-0.174$, $+0.007$] & no \\
Claude Sonnet 4.6 -- Regex A (stoplist)      & Recall & $-0.104$ & [$-0.214$, $+0.019$] & no \\
QWEN3.6-36B -- Regex A (stoplist)            & Recall & $-0.143$ & [$-0.266$, $-0.029$] & yes \\
GPT-5.2 -- Regex A (stoplist)                & Recall & $-0.182$ & [$-0.311$, $-0.064$] & yes \\
Claude Sonnet 4.6 -- Gemma4:31B              & Recall & $-0.253$ & [$-0.376$, $-0.146$] & yes \\
QWEN3.6-36B -- Gemma4:31B                    & Recall & $-0.292$ & [$-0.376$, $-0.205$] & yes \\
GPT-5.2 -- Gemma4:31B                        & Recall & $-0.331$ & [$-0.441$, $-0.215$] & yes \\
\bottomrule
\end{longtable}
}

\newpage
\section*{Appendix D: Properties of metadata}
\begin{table}[ht]
\centering
\small
\caption{Descriptive statistics of the YouTube metadata}
\label{tab:metadata_descriptives}
\begin{tabular}{lrrrrrr}
\toprule
Feature & Mean & SD & Median [IQR] \\
\midrule
Title length (characters) & 58.1 & 20.5 & 57.0 [43.0--73.0] \\
Title length (words) & 9.9 & 3.6 & 10.0 [7.0--12.0] \\
Description length (characters) & 1,196 & 1,084 & 901.0 [390.5--1,642] \\
Description length (words) & 166.5 & 159.2 & 121.0 [49.2--225.0] \\
URLs in description & 4.7 & 5.9 & 3.0 [0.0--7.0] \\
Hashtags in description & 2.4 & 5.4 & 0.0 [0.0--3.0] \\
Duration (minutes) & 23.1 & 54.6 & 5.5 [2.9--16.5] \\
View count (thousands) & 15,852 & 82,810 & 339.8 [8.9--3,508] \\
Upload year & 2021 & 4.6 & 2022 [2018--2025] \\
\bottomrule
\end{tabular}
\end{table}

\section*{Appendix E: Threshold-free performance}
ROC-AUC and PR-AUC (average precision) computed on the keep score of each LLM, without a threshold
($n=1047$, KEEP as positive class, 95\% cluster-bootstrap CIs over search queries, 1000 iterations).
The no-LLM baselines only produce binary decisions and therefore have no threshold-free score.
The PR-AUC of a random classifier equals the KEEP prevalence (0.147).
\begin{table}[ht]
\centering
\small
\caption{Threshold-free classification performance at metadata level ($n=1047$).}
\label{tab:threshold_free}
\begin{tabular}{lcccc}
\toprule
\textbf{Model} & \textbf{ROC-AUC} & \textbf{95\% CI} & \textbf{PR-AUC} & \textbf{95\% CI} \\
\midrule
QWEN3.6-36B        & \textbf{0.819} & [0.780, 0.854] & \textbf{0.370} & [0.283, 0.462] \\
Gemma4:31B         & 0.730 & [0.690, 0.776] & 0.261 & [0.200, 0.333] \\
GPT-5.2            & 0.728 & [0.678, 0.775] & 0.292 & [0.216, 0.389] \\
Claude Sonnet 4.6  & 0.760 & [0.710, 0.808] & 0.314 & [0.237, 0.406] \\
\bottomrule
\end{tabular}
\end{table}

\section*{Appendix F: Tuned versus fixed threshold}
Performance of each LLM with the fixed threshold $t=0.5$, which corresponds to the model's own KEEP/DISCARD decision,
and with the tuned threshold $t$ from the GroupKFold procedure (Section 3), at metadata level ($n=1047$).
$\Delta$ is tuned minus fixed, with 95\% CIs from a paired cluster bootstrap over search queries (1000 iterations).
\begin{table}[ht]
\centering
\small
\caption{Tuned versus fixed threshold ($t=0.5$), metadata level ($n=1047$).}
\label{tab:threshold_comparison}
\begin{tabular}{llccrcc}
\toprule
\textbf{Model} & \textbf{Metric} & \textbf{$t=0.5$} & \textbf{Tuned $t$} & \textbf{$\Delta$} & \textbf{95\% CI} & \textbf{Sig.} \\
\midrule
QWEN3.6-36B        & Precision & 0.298 & 0.354 & $+0.055$ & [$+0.019$, $+0.096$] & yes \\
                   & Recall & 0.851 & 0.565 & $-0.286$ & [$-0.358$, $-0.208$] & yes \\
                   & F1 & 0.442 & 0.435 & $-0.007$ & [$-0.055$, $+0.041$] & no \\
\midrule
Gemma4:31B         & Precision & 0.232 & 0.261 & $+0.030$ & [$+0.013$, $+0.048$] & yes \\
                   & Recall & 0.961 & 0.857 & $-0.104$ & [$-0.168$, $-0.045$] & yes \\
                   & F1 & 0.373 & 0.401 & $+0.027$ & [$+0.003$, $+0.051$] & yes \\
\midrule
GPT-5.2            & Precision & 0.199 & 0.285 & $+0.086$ & [$+0.041$, $+0.129$] & yes \\
                   & Recall & 0.838 & 0.526 & $-0.312$ & [$-0.409$, $-0.225$] & yes \\
                   & F1 & 0.322 & 0.370 & $+0.048$ & [$-0.001$, $+0.102$] & no \\
\midrule
Claude Sonnet 4.6  & Precision & 0.199 & 0.286 & $+0.087$ & [$+0.048$, $+0.130$] & yes \\
                   & Recall & 0.935 & 0.604 & $-0.331$ & [$-0.419$, $-0.234$] & yes \\
                   & F1 & 0.328 & 0.388 & $+0.060$ & [$+0.006$, $+0.103$] & yes \\
\bottomrule
\end{tabular}
\end{table}

\end{document}